# A Complexity Analysis of Statistical Learning Algorithms

Mark A. Kon, Boston University

**Abstract**

We apply information-based complexity analysis to support vector machine (SVM) algorithms, with the goal of a comprehensive continuous algorithmic analysis of such algorithms. This involves complexity measures in which some higher order operations (e.g., certain optimizations) are considered primitive for the purposes of measuring complexity. We consider classes $\mathcal{N}$ of information operators and $\Phi$ of algorithms made up of scaled families $\mathcal{N}_n \subset \mathcal{N}$ and $\Phi_k \subset \Phi$, and investigate the utility of scaling the complexities $n$ and $k$ to minimize error. We look at the division of statistical learning into information and algorithmic components, at the complexities of each, and at applications to support vector machine (SVM) and more general machine learning algorithms. We give applications to SVM algorithms graded into linear and higher order components, and give an example in biomedical informatics.

## 1. Introduction

This paper examines some applications of continuous complexity and information-based complexity theory (IBC; Traub, Woźniakowski and Wasilkowski, 1988, Traub and Werschulz, 1998) to statistical learning theory (SLT; Vapnik, 1998, 2000) and support vector machine (SVM) algorithms. One goal of this is to develop a more algorithmic formulation of SLT and SVM, and to study a context in which the complexity of these algorithms and other related ones can be analyzed. This algorithmic formulation of statistical learning approaches involves the treatment of some higher order computations (e.g., optimizations) as primitive operations for purposes of complexity calculations, and the use of graded families of information operators and algorithms (see below).

An element of this work is based on the fact that currently, higher-order computations are often rapid and effectively transparent to the user. For example, since SVM quadratic programming algorithms are computationally efficient, one can for some purposes view the operation of finding an optimal linear partition of two sets as primitive (and of unit complexity) in calculations. As a different example, in some cases information error estimation for an information operator $N$ (for the purpose of optimizing within a class $\mathcal{N}$) can be treated as primitive, since rapid cross-validation methods now make it possible to estimate such errors efficiently. Using such an approach, some algorithms, e.g. some classifications in computational biology (see below), can be studied more usefully.

Our algorithmic study of SLT approximation is formulated in terms of IBC, and in particular separates the solution process into information and algorithmic components. We will apply statistical learning approaches to grade information and algorithms by complexity (Vapnik, 1998, 2000) for minimization of error. For collections $\mathcal{N}$ of information operators and $\Phi$ of algorithms, since optimization in both spaces is needed, it can be useful to restrict one of these spaces to a small (low dimensional) one, so optimization in that space can be viewed as primitive or of low complexity, and so study can be focused on the other (larger) space.

We note that other related algorithmic analyses of continuous computation have also been studied in recent years; see Blum, et al. (1998) and Braverman and Cook (2006).

## 1.1 Continuous complexity notions in learning theory

Let $F$ be a space of problem elements, $S : F \to G$ be a solution operator, and $N : F \to \mathbb{R}^n$ an information operator, with $y = Nf$ a vector containing all available information about the problem $f \in F$. Thus the problem element $f \in F$ (e.g., a PDE with given boundary conditions) corresponds to an exact problem solution $Sf \in G$ (the PDE solution). Note in applications that $f$ may be unknown, with information about $f$ in the form $Nf \in \mathbb{R}^n$.

For our learning theory applications we will assume $\rho = f \in F$ is an (unknown) input-output relation in the space $F = \mathcal{P}$ of all probability distributions on an input-output space $\mathbf{X} \times \mathbb{R} = \{(\mathbf{x}, y) : \mathbf{x} \in \mathbf{X}, y \in \mathbb{R}\}$. (The notations $\rho$ as well as $f$ will be used, since they are standard in different contexts). We seek the solution $S\rho = g(\mathbf{x}) \in G$, a function $g : \mathbf{X} \to \mathbb{R}$ that is the best approximation in function class $G$ of the probabilistic relation $\rho$ between $\mathbf{x}$ and $y$, by some error criterion. An example of such a criterion is the loss

$$R(g) = E_\rho(L(g(\mathbf{x}), y)) \equiv \int_{\mathbb{R}^d \times \mathbb{R}} L(g(\mathbf{x}), y) d\rho(\mathbf{x}, y),$$

where $L(g(\mathbf{x}), y)$ measures a distance between $y$ and its approximation $g(\mathbf{x})$. The set $G$ of allowed relations between $\mathbf{x}$ and $y$ is the *hypothesis space.*

**SVM algorithms:** In the case of SVM (Vapnik, 1995, 1998) we deal with a classification problem, and the output is restricted to $y \in \mathbb{B} = \{1, -1\} \subset \mathbb{R}$. We incorporate this case by restricting the support of $\rho(\mathbf{x}, y)$ to $y = \pm 1$. The goal is finding $S\rho = g(\mathbf{x}) \in G$, where $g(\mathbf{x})$ best approximates the $\rho$-induced $\mathbf{x}$-$y$ relationship, by minimizing the average difference between $\text{sgn}(g(\mathbf{x}))$ (sign of $g(\mathbf{x})$) and $y$, as measured by letting $L(g, y) = (1 - gy)_+$ above, with

$$a_+ = \begin{cases} a & \text{if } a \geq 0 \\ 0 & \text{otherwise} \end{cases}.$$

**Information:** The goal is to find an algorithm $\phi \in \Phi$ (with $\Phi$ an algorithm class) which best uses information $N\rho$ to approximate $S\rho$ with $\phi(N\rho)$. The (random) information in this case takes the form $N\rho = \{\mathbf{z}_i = (\mathbf{x}_i, y_i)\}_{i=1}^n$, where $\mathbf{z}_i \in \mathbf{X} \times \mathbb{R}$ are chosen independent and identically distributed (iid) according to $\rho$. The space $\mathbf{X}$ is then the *feature space.*

**Algorithm optimization:** The standard SVM algorithm restricts the class of $\phi$ to $\Phi_1 \subset \Phi$, with $\Phi_1$ the set of $\phi_1$ which map into the class of affine functions $\phi_1(N\rho) = g(\mathbf{x}) = \mathbf{w} \cdot \mathbf{x} + b$, so the range $\text{ran}(\phi) \subset G_1$. In some cases we can treat the computation of $g = \phi_1(\rho)$ from data $N\rho$ as primitive (e.g., using standard quadratic programming), and focus on finding optimal information $N \in \mathcal{N}$ which minimizes error and has small cardinality. Information optimization is important, e.g., in computational biology applications, where dim $\mathbf{X}$ may be on the order of $10^5$ or more.

We then need the right set $\mathcal{N}$ of information operators and wish to minimize risk (error) over $N \in \mathcal{N}$. The focus on error then moves from the algorithm space $\Phi_1$ to $\mathcal{N}$. In some cases the optimization on $\mathcal{N}$ may focus on dimension reduction of the feature space, for example choosing $N(\rho) = N_P(\rho) = \{(P\mathbf{x}_i, y_i)\}$, where $P$ is a projection in $\mathbf{X}$. This can give a choice of coordinates in $\mathbf{X}$ "relevant" to predicting $y$ from $\mathbf{x}$. Such optimizations can be interesting in their own rights; for example, improved coordinate sets might be obtained using genetic algorithms or simulated annealing, involving adiabatic replacement of "worse" coordinates $x_i$ by better ones $x_i'$, as well as larger searches for subspaces of $\mathbf{X}$ optimized for separation of positive and negative $y$ values.

## 1.2 Grading of information and algorithmic complexities

We consider complexity-based scalings between increasing families $\mathcal{N}_n$ and $\Phi_k$ of information and algorithm spaces. Given an error criterion $|R[\phi(N\rho)] - R[S\rho]|$ (section 2) for the difference between the exact solution $S\rho$ and its approximation $\phi(N\rho)$, assume there is an optimal pair $(N, \phi) \in \mathcal{N} \times \Phi$ giving the best approximation, with $\mathcal{N} \supset \bigcup_n \mathcal{N}_n$ and $\Phi \supset \bigcup_k \Phi_k$ fixed. Analytic and computational resources may be limited, and there is risk of overfitting if the range $G$ of the class $\Phi$ is too large. Thus in practice it is useful to search for restricted optima, with the optimization done over sub-classes of $\mathcal{N}$ or $\Phi$ or both. Thus one can restrict to a very small set of (practical) algorithms, e.g., for SVM to $\Phi_1 \subset \Phi$ as above, with $\Phi_1$ mapping into affine $g(\mathbf{x})$ only. It can also be useful (as in SLT applications), to scale algorithmic and information complexity, i.e., the complexity of the (range of the) working subset $\Phi_k \subset \Phi$ with information complexity of a subset $\mathcal{N}_n \subset \mathcal{N}$.

For example, such a scaling is useful in choosing between linear and nonlinear regression. To guess an unknown smooth scalar function $f \in F$ (with $F$ now a function class on $\mathbb{R}$) from data $Nf = (f(x_1), \ldots, f(x_n))$, one can adjust the class of the algorithm $\phi$ (linear versus higher order regression) for determining the best fit to information of cardinality $n$. When computational methods first became widely available for inference, even large databases were still fit to linear $f$, so $\phi$ gave the closest linear approximation to $f$. However, unless a priori information indicates $f$ is linear, this is not necessarily a good choice. In particular, if $n$ is large enough, adding a quadratic component to $f$ will still yield sufficiently few free parameters to fit the data. Thus in principle algorithmic complexity should be scaled with information complexity (Vapnik, 1998, 2000), i.e., the number of parameters (monotonic in $k$) should be scaled with the amount of data (monotonic in $n$). Thus we adjust size of the hypothesis space $G_k$ to data cardinality $n$.

# 2. Notation and definitions

## 2.1 Information, algorithm, error

We now give the assumptions more precisely. Let $F$ be a space (e.g., a function space or space of probability distributions) containing an unknown element $f \in F$, describing the state of a real system. We are interested in an element $g = Sf \in G$ which solves a problem whose input is $f$. Here $S : F \to G$ denotes a linear or nonlinear *solution* operator. We assume there is an information operator $N : F \to \mathbb{R}^n$ (the *information space*), chosen from a family $\mathcal{N}$ of possible information operators. We assume a family $\Phi$ of allowed *algorithms* $\phi : \mathbb{R}^n \to G$, and wish to choose a $\phi \in \Phi$ and $N \in \mathcal{N}$ for which the approximation $\phi(N(f)) \approx S(f)$ is optimal in a given error measure. Informally, we seek $N \in \mathcal{N}$ and $\phi \in \Phi$ so the diagram commutes "maximally"; see Traub, Woźniakowski and Wasilkowski (1988) and Traub and Werschulz (1998) for a further description of this problem in general. We assume a class of "primitive" operations, assumed to have unit complexity, whose cardinality in the problem solution will measure complexity, and focus the analysis on the particular choice of subspaces of $\mathcal{N}$ and $\Phi$.

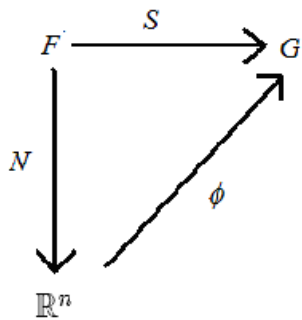

**SVM learning:** To consider the example of SVM learning, we assume a more precise SLT model, in which the unknown $f = \rho(\mathbf{x}, y)$ represents a relationship between $\mathbf{x}$ and $y$. Here $\rho \in F = \mathcal{P}$, the space of probability distributions (all unit Borel measures, not necessarily with density functions) on $\mathbf{X} \times \mathbb{R}$. $\mathbf{X}$ is the *feature space*. For simplicity we assume $\mathbf{X} = \mathbb{R}^d$, and we can restrict to subsets of $\mathbb{R}^d$ by allowing $\rho$ to have compact support if we choose. Thus $\mathbf{x}$ and $y$ are input and output, with $\rho$ determining their relationship. Information about $\rho$ is provided in Monte Carlo fashion as examples $N\rho = \{\mathbf{z}_i = (\mathbf{x}_i, y_i)\}_{i=1}^n$, with $\mathbf{z}_i \in \mathbf{X} \times \mathbb{R}$ chosen iid from $\rho$.

We seek $g(\mathbf{x}) : \mathbf{X} \to \mathbb{R}$ such that given $\mathbf{x}$, and the choice $y = \operatorname{sgn} g(\mathbf{x})$, the graph $\{(\mathbf{x}, g(\mathbf{x}))) : \mathbf{x} \in \mathbb{R}^d\}$, with $\mathbf{x}$ random and distributed according to (the marginal of) $\rho$ on $\mathbb{R}^d$, best approximates the distribution $\rho(\mathbf{x}, y)$. More precisely, let the local loss $L(g(\mathbf{x}), y)$ represent a distance between $y$ and its approximation $g(\mathbf{x})$. Two choices might be

$$L = L_1(\mathbf{z}) = \|g(\mathbf{x}) - y\|_2,$$

and (in the case where $y_i = \pm 1$)

$$L_2(\mathbf{z}) = (1 - yg(\mathbf{x}))_+,$$

which is the hinge loss in the classical SVM; we assume $L \geq 0$.

Define the risk $R$ by

$$R(g) \equiv R_\rho(g) \equiv \int_{\mathbb{R}^d \times \mathbb{R}} L(g(\mathbf{x}), y) d\rho(\mathbf{x}, y) \equiv E_\rho(L(g(\mathbf{x}), y)). \quad (2.1)$$

We let $R = R_\rho$ and $E = E_\rho$ unless otherwise specified.

Define the $S$ taking $\rho$ to the best approximation $g_0$ by

$$S\rho = g_0 = \arg\min_{g \in G} R(g)$$

(assuming the minimum exists and is unique), so that $S\rho$ minimizes (and takes to 0) the deviation $R(g) - R(g_0)$ of the risk of $g$ from the optimal risk $R(g_0)$. We seek an algorithm $\phi : N\rho \to G$ which best approximates $S$. We select $\phi$ from a "small" more easily computable class of algorithms $\Phi_1$ whose ranges are in the subset $G_1 \subset G$, with $G_1$ in this case consisting of affine separator functions $g(\mathbf{x}) = \mathbf{w} \cdot \mathbf{x} + b$.

**Definition of error:** We formally define the error between our approximation $g \in G_1$ and the optimal $g_0 \in G$ by

$$e(g, g_0) = R(g) - R(g_0) \quad (2.2)$$

(Vapnik, 1998, 2000).

In many cases an error measure like $e$, at least for a finite dimensional hypothesis space $G$, is equivalent to "standard" error measures (at least when one of the arguments is the optimal $g_0$). Indeed assume the risk functional $R(g)$ for $g \in G$ is a twice differentiable function of $g \in G$ (assumed finite dimensional). For a unique minimum $g_0$, the Hessian matrix $H(g_0)$ must be positive. If in addition it is positive definite, it follows that for any metric defined by a norm $\|\cdot\|_G$ on $G$, there are $c_1, c_2 > 0$ such that

$$c_1 \|g - g_0\|_G^2 \leq R(g) - R(g_0) \leq c_2 \|g - g_0\|_G^2. \tag{2.3}$$

Then up to a constant the error in (2.2) can be replaced by $e(g, g_0) = \|g - g_0\|_G^2$.

Further, for any finite or infinite dimensional choice of $G$, if we use an $L^2$-type measure of risk,

$$R(g) = \int_{\mathbb{R}^{d+1}} (g(\mathbf{x}) - y)^2 d\rho(\mathbf{x}, y)$$

we have an equality in (2.3). Namely, since $g_0(\mathbf{x}) = E_\rho(y|\mathbf{x})$,

$$R(g) - R(g_0) = E\big[(g(\mathbf{x}) - y)^2 - (g_0(\mathbf{x}) - y)^2\big]$$

$$= E_\mathbf{x}\big(E\big[(g(\mathbf{x}) - y)^2 - (g_0(\mathbf{x}) - y)^2 \,\big|\, \mathbf{x}\big]\big)$$

$$= E_\mathbf{x}\big((g^2 - g_0^2 + 2E(y|\mathbf{x})(g_0 - g))\big)$$

$$= E_\mathbf{x}\big((g^2 - g_0^2 + 2g_0(g_0 - g))\big)$$

$$= \|g - g_0\|_2^2, \tag{2.4}$$

so the error reduces to an $L^2$ one. Above $E_\mathbf{x}$ is marginal expectation with respect to $\mathbf{x}$.

## 2.2 Information and algorithmic error

Assume as above there is an optimal $g_0 \in G$, and we wish to approximate $g_0$ in $G_1 \subset G$. More generally assume a scale of increasing spaces $G_k \subset G$, and that we seek $g_k \in G_k$ as the best approximation of $g_0$ in $G_k$. Let $\Phi_k$ denote the set of all algorithms on the information space $\mathbb{R}^n$ which map into $G_k$, i.e., $\phi_k : \mathbb{R}^n \to G_k$ for $\phi_k \in \Phi_k$. That is the algorithm space $\Phi_k$ consists of *all* maps from $\mathbb{R}^{n+1}$ to $G_k$ (we always restrict functions to be measurable); thus $\Phi_k$ depends on $G_k$ only. We seek a $\phi_k \in \Phi_k$ so that $\phi_k(N\rho) \equiv \widehat{g}_k \in G_k$ is an approximation of $g_k$ and thus of $g_0$.

**Error definitions:** There are two sources of error in the approximation of $g_0$ by $\widehat{g}_k$. First for $\phi_k : \mathbb{R}^d \to G_k \subset G$, there is an algorithmic (approximation) error determined entirely by the choice of approximation space $G_k$ (equivalently $\Phi_k$), given by

$$e_{\text{alg}}(k) = \inf_{g \in G_k} e(g, g_0) = \inf_{g \in G_k} R(g) - R(g_0) = R(g_k) - R(g_0).$$

Here

$$g_k \equiv \arg\inf_{g \in G_k} (R(g) - R(g_0))$$

represents any choice which minimizes $R(g) - R(g_0)$, assuming such a minimizer exists (we assume existence of a minimizer, but if it does not exist the definition can be modified with approximate minima or minimizing sequences). Note that this and the other errors below are *local* errors, since they depend on the probability distribution $\rho$.

Second there is information (estimation) error,

$$e_{\text{inf}}(n,k) = \inf_{\phi \in \Phi_k} E(R[\phi(N_n\rho)] - R[g_k])$$

$$= \inf_{\phi \in \Phi_k} E(R[\phi(\mathbf{Z})] - R[g_k]),$$

where $N_n\rho = \{\mathbf{z}_i = (\mathbf{x}_i, y_i)\}_{i=1}^n \equiv \mathbf{Z}$ are chosen iid according to $\rho$, and the above expectation $E = E_{\mathbf{Z}}$ is over $\mathbf{Z}$.

Thus $e_{\text{inf}}(n,k)$ is the error if we use the best possible algorithm $\phi_k \in \Phi_k$ into $G_k$, using random information $N\rho = \mathbf{Z}$ limited to cardinality $n$. The error is averaged over the choice of Monte Carlo information $\mathbf{Z}$ under $\rho$. On the other hand $e_{\text{alg}}(k)$ is the error with full information, if the algorithm is restricted to the class $\Phi_k$ of all algorithms mapping into $G_k$. This error is characterized in SLT (Vapnik, 1998) as approximation error. See Kon and Plaskota (2000) for a study of relationships between these two types of errors in the context of neural network algorithms. Approximation error in this context has also been studied in Smale and Zhou (2003) and in Kon and Raphael (2006).

For convenience assume a minimizer

$$\phi_k = \arg\inf_{\phi \in \Phi_k} E(R[\phi(\mathbf{Z})] - R[g_k])$$

exists; otherwise we can again use minimizing sequences for the following definitions. Letting $\widehat{g}_k \equiv \phi_k(\mathbf{Z})$ be the algorithmic estimate of $g_0$, we define the full error

$$e(n,k) = E(R[\phi_k(\mathbf{Z})] - R[g_0])$$

$$= E(R(\widehat{g}_k)) - R(g_k) + R(g_k) - R(g_0) \quad (2.5)$$

$$= e_{\inf}(n,k) + e_{\mathrm{alg}}(k).$$

Thus by using effectively squared errors in our definitions of $e_{\inf}$ and $e_{\mathrm{alg}}$ (as in (2.4)), with our error $e(n,k)$ defined in terms of risks as above, have an exact equality (2.5), as opposed to bounds of the type in Kon and Plaskota (2000).

**Complexity:** We correspondingly separate the complexity of approximation into two parts, *information complexity* $n$ and *algorithmic complexity* $k$, the inverses of the functions $e_{\inf}(n,k)$ (information error) and $e_{\mathrm{alg}}(k)$ (algorithmic error). We will assume that the unit of complexity is normalized so that the cost of each additional information operation (obtaining of a data point $\mathbf{z}_i = (\mathbf{x}_i, y_i)$) is 1. We then define information complexity by

$$\mathrm{comp}_{\inf}(\epsilon, k) = \inf\{n : e_{\inf}(n,k) \leq \epsilon\}.$$

To define algorithmic complexity, we define for a $\phi : \mathbb{R}^n \to G$ (where $G$ contains $\bigcup_k G_k$) the algorithmic error

$$e(n,\phi) = E[R(\phi(N\rho))] - R(S\rho) = E[R(\phi(\mathbf{Z}))] - R(g_0).$$

We define $C(n,\phi)$ to be the complexity of the computation of $\phi$ from given information $N_n\rho$ (again in units where one information operation has cost 1).

Now define the algorithmic complexity as

$$\mathrm{comp}(\epsilon, n, k) = \inf_{\phi \in \Phi_k} \{C(n,\phi) : e(n,\phi) \leq \epsilon\}. \quad (2.6)$$

The full $\epsilon$-complexity of approximation in $G$ is defined as

$$\text{comp}(\epsilon) = \inf_{n,k,\phi \in \Phi_k} \{n + C(n,\phi) : e(n,\phi) \leq \epsilon\}. \tag{2.7}$$

This defines the best information cardinality $n$ and algorithmic complexity level $k$ to obtain an approximation of the optimal $g \in G$ within error $\epsilon$. As mentioned above, we generally will want to choose from a graded family of algorithm classes $\Phi_k$ whose complexity (which depends on the range $G_k$ of the family $\Phi_k$) will scale with the information cardinality $n$ in order to optimize (2.7). A natural question is how such a scaling should go, which is considered in the following discussion of SVM.

## 3. An example in SVM

We now restrict to a more specific application, the statistical learning theory (SLT) formulation of support vector machines, involving SLT and its algorithmic IBC formulation.

Again let $\mathcal{P}$ be the space of probability distributions on $\mathbb{R}^{d+1}$, and $G$ a function space on $\mathbb{R}^d$. As before specify the map $S : F = \mathcal{P} \to G$ which takes a probability distribution $\rho \in \mathcal{P}$ to its functional best approximation $S\rho = g \in G$, with the goal of estimating $y \approx g(\mathbf{x})$. As above, we define $g = g_0 \in G$ to minimize the risk

$$R(g) \equiv E_\rho[L(g(\mathbf{x}), y)] = \int_{\mathbb{R}^{d+1}} L(g(\mathbf{x}), y) \, d\rho(\mathbf{x}, y).$$

We now assume $y$ is restricted to the values $\mathbb{B} = \{\pm 1\}$ through concentration of the measure $\rho(\mathbf{x}, y)$ on $y = \pm 1$, and that we are given the data set $N\rho = \mathbf{Z} = \{\mathbf{z}_i = (\mathbf{x}_i, y_i)\}_{i=1}^n$ (training data) consisting of random information derived from $\rho \in F \equiv \mathcal{P}$, with $\rho$ unknown. $\mathbf{Z}$ represents example classifications $y_i$ (positive or negative) of data $\{\mathbf{x}_i\}_{i=1}^n$. The risk-minimizing $g$ is intended to generalize the examples $\mathbf{Z}$, and when $g(\mathbf{x}) \geq 0$, a new $\mathbf{x}$ is classified as positive, and otherwise negative. We now restrict the set of classification functions $G$ to $G_1 \subset G$ (consisting of affine functions on $\mathbb{R}^d$) as in SVM, to simplify estimation.

**Empirical risk:** Let

$$g_1 = \arg\inf_{g \in G_1}(R(g) - R(g_0)) = \arg\inf_{g \in G_1} R(g)$$

be the closest element in $G_1$ to the optimal $g_0 \in G$. To formulate the choice of algorithm $\phi_1 : \mathbf{Z} \to \widehat{g}_1$ estimating $g_1$, we define the *empirical probability distribution* $\widehat{\rho}$ to be our estimate of $\rho$ given data $\mathbf{Z} = (\mathbf{z}_1, \ldots, \mathbf{z}_n)$. It is defined by $\widehat{\rho} = \frac{1}{n}\sum_i \delta_{\mathbf{z}_i}(\mathbf{z})$, where $\mathbf{z} = (\mathbf{x}, y)$ and $\delta_{\mathbf{z}}$ is the point mass at $\mathbf{z} \in \mathbb{R}^{d+1}$.

The *empirical risk* of any $g$ is the corresponding estimate of true risk $R(g) \equiv R_\rho(g)$, i.e.,

$$R_{\widehat{\rho}}(g) = E_{\widehat{\rho}}(L(g(\mathbf{x}), y)) = \int_{\mathbb{R}^{d+1}} L(g(\mathbf{x}), y) d\widehat{\rho}(\mathbf{x}, y)$$

$$= \frac{1}{n}\sum_{i=1}^n L(g(\mathbf{x}_i), y_i). \tag{3.1}$$

Let $\widehat{g}_1 = \arg\inf_{g \in G_1} R_{\widehat{\rho}}(g)$ be the empirical risk minimizer, again assuming that it exists, which is case under weak hypotheses. The minimizer $\widehat{g}_1$ defines the "$L$-regression" separator, an approximation of the optimal separator $g_1$, (the true minimizer of $R_\rho$ in $G_1$). Define the algorithm $\phi_1 \in \Phi_1$ (the set of all maps from $\mathbb{R}^d$ to $G_1$) by $\phi_1(\mathbf{Z}) = \widehat{g}_1$.

Thus the hypothesis space $G_1 \subset G$ consists of affine $g_1(\mathbf{x}) = \mathbf{w} \cdot \mathbf{x} + b$ on $\mathbb{R}^d$, and restriction to $G_1$ is done by limiting $\Phi_1$ to algorithms with range in $G_1$. We will discuss in section 6 the consideration of a larger scale of SVM algorithm spaces $\Phi_k = \Phi_{k(n)}$ forming a nested family (differing in their ranges $G_{k(n)}$) which scale with cardinality $n$ of information, though for now we fix $k = 1$ and $\Phi = \Phi_1$.

## 4. SVM: Convergence rates of empirical risks

### 4.1 Risk and VC dimension

We now consider some complexity bounds on SVM algorithms. We define the local error of the algorithm (which depends on $\rho \in F$) to be

$$e(N, \phi) \equiv E[R(\phi(N\rho)) - R(g_0)],$$

where $g_0 = S\rho$ is the true minimizer of risk, assumed to exist in the class $G$ of all functions on $\mathbb{R}^d$. Recall $N\rho = \mathbf{Z} = (\mathbf{z}_1, \mathbf{z}_2, \ldots, \mathbf{z}_n)$, with $\mathbf{z}_i = (\mathbf{x}_i, y_i) \in \mathbb{R}^{d+1}$ iid and chosen according to $\rho$.

To bound information error as the cardinality $n \to \infty$, there are several results in continuous complexity and SLT which are useful here. Letting

$$\|L(g(\mathbf{x}), y)\|_p = \left(\int_{\mathbb{R}^{d+1}} L(g(\mathbf{x}), y)^p \, d\rho(\mathbf{x}, y)\right)^{1/p} = E(L(g(\mathbf{x}), y)^p)^{1/p},$$

first we have an error bound based on results of Vapnik (1998, 2000). We will define for a given loss function $L(g(\mathbf{x}), y) = V(\mathbf{x}, y)$ its VC dimension, defined for the family $H = \{L(g(\mathbf{x}), y)\}_{g \in G}$. For any family of functions, this is defined by

**Definition 4.1:** A family $D$ of real-valued functions on a space $Z$ is said to *separate* a set of points $A = \{\mathbf{z}_i\}_{i=1}^n \subset Z$ if for every subset $A_1 \subset A$, there exists a $d \in D$ and $\alpha \in \mathbb{R}$ such that $d(\mathbf{z}) - \alpha > 0$ if and only if $\mathbf{z} \in A_1$.

**Definition 4.2:** The *VC dimension* of a family $D$ of functions on the space $Z$ is the cardinality of the largest set of points $A \subset Z$ which is separated by $D$. If this cardinality is unbounded then the VC dimension of $D$ is infinite.

## 4.2 Error estimates

We now discuss some asymptotic error estimates independent of the initial distribution $\rho(\mathbf{z}) \in F$. As above $G_1$ is the set of affine functions on $\mathbb{R}^d$. Henceforth let $p > 2$, and define

$$\tau = \sup_{g \in G_1} \frac{\|L(g(\mathbf{x}), y))\|_p}{\|L(g(\mathbf{x}), y)\|_1}, \tag{4.1}$$

and

$$a(p) = \frac{1}{2^{1/p}} \left( \frac{p-1}{p-2} \right)^{\frac{p-1}{p}}. \tag{4.2}$$

Define for any probability $\delta > 0$

$$\mathcal{E} \equiv 4 \frac{h(\ln \frac{2n}{h} + 1) - \ln(\frac{\delta}{8})}{n},$$

where $h$ is the VC dimension of the set of functions $D = \{L(g(\mathbf{x}), y)\}_{g \in G_1}$ on $\mathbb{R}^{d+1}$.

As above, we assume Monte Carlo information about the unknown relationship $\rho(\mathbf{x}, y)$ in the form $N\rho = \mathbf{Z} = \{\mathbf{z}_i\}_{i=1}^n$, with $\mathbf{z}_i = (\mathbf{x}_i, y_i)$ iid from $\rho$. We assume a minimizer $\widehat{g}_1$ of the empirical risk $R_{\widehat{\rho}}(g)$ exists. Then we have (see Vapnik, 2000, §3.7) a result which only depends on the VC dimension $h$ of $D$:

**Theorem 4.1 (Vapnik):** *For any non-negative loss L, we have with probability at least $1 - \delta$, that for random information $N\rho = \{(\mathbf{x}_i, y_i)\}_{i=1}^n$,*

$$\frac{R(\widehat{g}_1) - \inf_{g \in G_1} R(g)}{\inf_{g \in G_1} R(g)} \leq \frac{\tau a(p) \sqrt{\mathcal{E}}}{\left(1 - \tau a(p) \sqrt{\mathcal{E}}\right)_+} + O\left(\frac{1}{n}\right). \tag{4.3}$$

*where $\widehat{g}_1$ is a minimizer in $G_1$ of empirical risk $R_{\widehat{\rho}}(g)$.*

This gives a $\delta$-PAC (probably approximately correct, with probability greater than $1 - \delta$) bound on the SVM error. The formulation in Vapnik, 2000 is stated equivalently in terms of the probability parameter $\eta = \delta/2$. As mentioned earlier the error on the left side can generally be expressed in terms of a norm error on the finite dimensional hypothesis space $G_1$. Since $\tau$ depends on the unknown $\rho \in F$, the above error is local in $F$. The $\tau$ dependence can be eliminated if we assume, for example, that the loss $L(g, y)$ is bounded as a function of $g$ and $y$ (Vapnik, 2000, §3.7). We note that the $O\left(\frac{1}{n}\right)$ term above is uniform in the choice of $\rho$.

## 4.3 Complexity estimates

We now estimate information complexity of the standard SVM algorithm by inverting (4.3). We define the ($\delta$-PAC) $\epsilon$-information complexity $n$ of finding the risk-minimizing $g_1$ by:

$$n = \text{comp}_{\inf}(\epsilon),$$

$$= \inf_{n'}\{|R_\rho(\widehat{g}_1(n')) - R_\rho(g_1)| < \epsilon \text{ with probability at least } 1 - \delta\}.$$

where $g_1 = \arg\inf_{g \in G_1}(R(g))$ and

$$\widehat{g}_1(n') = \arg\inf_{g \in G_1} R_{\widehat{\rho}}(g))$$

with $\widehat{\rho} = \frac{1}{n'}\sum_{i=1}^{n'}\delta_{\mathbf{z}_i}$ formed from information $N\rho = \{\mathbf{z}_i\}_{i=1}^{n'}$. Letting $J = \inf_{g \in G_1} R(g)$, we have from Theorem 4.1 (always with $\rho$-probability at least $1 - \delta$)

$$\epsilon \leq \frac{(J\tau a)\sqrt{\mathcal{E}}}{(1 - \tau a\sqrt{\mathcal{E}})_+} + O(1/n),$$

where $a = a(p)$ is as in (4.2). Note that by our definitions we should in fact have $n$ instead of $n-1$ on the right side; this change is however absorbed in the $O(1/n)$ term. Thus

$$\epsilon \leq b\frac{\sqrt{\frac{h\ln n + q}{n}}}{\left(1 - e\sqrt{\frac{h\ln n + q}{n}}\right)_+} + O(1/n)$$

$$= c\sqrt{\frac{\ln n}{n}} + k\frac{1}{\sqrt{n\ln n}} + O\left(\frac{1}{\sqrt{n}\ln^{3/2} n}\right),$$

defining

$$e = 2\tau a; \quad b = 2J\tau a; \quad q = h(1 + \ln\frac{2}{h}) - \ln\frac{\delta}{8}; \quad c = b\sqrt{h}; \quad k = \frac{bq}{2\sqrt{h}};$$

the above holds for $n$ sufficiently large that $1 - e\sqrt{\frac{h\ln n + q}{n}} > 0$.

To invert this for $n$, we invert the corresponding equality, replacing $\epsilon$ by $\epsilon' \geq \epsilon$ defined by

$$\epsilon' = c\sqrt{\frac{\ln n}{n}} + k\frac{1}{\sqrt{n\ln n}} + O\left(\frac{1}{\sqrt{n}\ln^{3/2} n}\right), \tag{4.4}$$

so after squaring and letting $\beta = \left(\frac{\epsilon'^2}{c^2}\right)$, $A = \frac{2k}{c} = \frac{q}{h}$,

$$\beta = \left(\frac{\ln n}{n} + A \cdot \frac{1}{n} + B(n)\right) \tag{4.5}$$

where $B(n) = O\left(\frac{1}{n\ln n}\right)$.

Defining $D = e^A$ and letting

$$\gamma = \frac{\beta}{D} = \frac{\epsilon'^2}{2(2J\tau a)^2 e\left(\frac{8}{\delta}\right)^{1/h}}; \quad m = Dn = \frac{2e}{h}\left(\frac{8}{\delta}\right)^{1/h} n,$$

we have

$$\gamma = \frac{\ln m}{m} + B(m), \tag{4.6}$$

where $B(m) = O\left(\frac{1}{m\ln m}\right)$.

We insert a solution of (4.6) of the form

$$m = \left(\frac{1}{\gamma}(-\ln\gamma) + r\right) \tag{4.7}$$

with the expectation that $r$ is of lower order than $\frac{1}{\gamma}(-\ln\gamma)$ as $\gamma \to 0$. To validate this and estimate $r$, let $H = \left(\frac{-\ln\gamma}{\gamma}\right)$. Using (4.7) in (4.6), (below $a_1$ is the first MacLaurin coefficient of $\frac{\ln(1+x)}{1+x}$ )

$$\gamma = \gamma + \gamma \frac{\ln(-\ln\gamma)}{-\ln\gamma} + r\frac{\gamma^2}{\ln^2\gamma}\ln\gamma - r\frac{\gamma^2}{\ln^2\gamma}\ln(-\ln\gamma) + r\frac{\gamma^2}{\ln^2\gamma} a_1$$

$$+ B + O\left(\gamma^3 \frac{r^2}{\ln^2\gamma}\right) \tag{4.8}$$

Note that

$$B = O\left(\frac{1}{m\ln m}\right) = o\left(\frac{\gamma}{-\ln\gamma}\right) = o\left(\gamma \frac{\ln(-\ln\gamma)}{-\ln\gamma}\right) \quad (\gamma \to 0), \tag{4.9}$$

since

$$\frac{\gamma}{-\ln\gamma} = \frac{\frac{\ln m}{m} + B}{-\ln\left(\frac{\ln m}{m} + B\right)} = \Omega\left(\frac{\ln m}{m \ln m}\right) = \Omega\left(\frac{1}{m}\right) \tag{4.10}$$

by (4.6), where for any function $f(m)$, by definition $\frac{\Omega(f(m))}{f(m)} \geq c > 0$ as $m \to \infty$.

Thus by (4.8) and (4.9), as $\gamma \to 0$,

$$0 = \gamma \frac{\ln(-\ln\gamma)}{-\ln\gamma}(1 + o(1)) + r\frac{\gamma^2}{\ln^2\gamma}\ln\gamma(1 + o(1)) \tag{4.11}$$

so

$$r = \frac{\ln(-\ln\gamma)}{\gamma}(1 + o(1)).$$

Thus by (4.7)

$$m = \frac{-\ln\gamma}{\gamma} + \frac{\ln(-\ln\gamma)}{\gamma} + o\left(\frac{\ln(-\ln\gamma)}{\gamma}\right).$$

or

$$n = \frac{1}{\beta}\left(\ln\left(\frac{D}{\beta}\right) + \ln\ln\left(\frac{D}{\beta}\right)(1 + o(1))\right)$$

$$= \frac{4J^2\tau^2 a^2 h}{\epsilon'^2}\left(2\ln\left(\frac{1}{\epsilon'}\right) + \ln\left(\ln\frac{1}{\epsilon'}\right)(1+o(1))\right). \tag{4.12}$$

We note more information is needed in the asymptotic expansion to determine uniform dependence on $\delta$. The equality in $\epsilon'$ above in (4.4) is easily replaced by the inequality in $\epsilon \leq \epsilon'$ again (since all functions are monotonic). Recalling that one information operation is a unit of information complexity, we have

**Theorem 4.2:** *Given an allowed probability $\delta$ of error $\epsilon > 0$, the information complexity of the $\delta$-PAC approximation for the support vector machine in $d$ dimensions is bounded by*

$$n \leq (d+2)(2J\tau a)^2 \left(2\frac{\ln\frac{1}{\epsilon}}{\epsilon^2} + \frac{\ln\ln\left(\frac{1}{\epsilon}\right)}{\epsilon^2}\right) + o\left(\frac{\ln\ln\left(\frac{1}{\epsilon}\right)}{\epsilon^2}\right) \tag{4.13}$$

*as* $\epsilon \to 0$, *where* $J = \inf_{g \in G_1} R(g)$ *denotes the minimal risk, and* $\tau = \sup_{g \in G_1} \frac{\|L(g(\mathbf{x}),y)\|_p}{\|L(g(\mathbf{x}),y)\|_1}$, $a = \frac{1}{2^{1/p}}\left(\frac{p-1}{p-2}\right)^{\frac{p-1}{p}}$ *for any* $p > 2$.

The theorem follows from (4.12) since the VC dimension $h$ of the space $G_1$ of affine functions on the feature space $\mathbf{X} = \mathbb{R}^d$ is $d+1$ (note that up to this point the argument is valid for a general loss $L$). Now letting $L(g(\mathbf{x}), y) = (1 - yg(\mathbf{x}))_+$, the VC dimension of this family of loss functions is determined by first noting $L(g(\mathbf{x}), y)$ is a monotonic function of $1 - yg(\mathbf{x})$, and hence its VC dimension is bounded by that of the latter. The VC dimension of $1 - yg(\mathbf{x})$ can be bounded by noticing that (since $g$ is affine) it is an affine combination of the functions $\{yx_i\}_{i=1}^d$ and $y$, forming (upon mapping $(\mathbf{x}, y)$ into $(y\mathbf{x}, y)$ as a new coordinate system) affine functions in a coordinate system of $d+1$ dimensions. In a $d+1$ dimensional coordinate system the set of affine functions has VC dimension $h$ bounded by $d+2$. Thus we replace the VC dimension $h$ by $d+2$ on the right side of (4.13) (see Vapnik, 1998, Korian and Sontag, 1997).

This gives us a $\delta$-probabilistic complexity estimate, giving a $\delta$-PAC complexity to first order. That is, with information of cardinality $n$ we can obtain an approximate solution to the SVM problem, with a (probably) small risk. Note as above that this is a *local* complexity (dependent on $\rho \in F$) if we do not assume the risk function $L$ is bounded.

## 4.4. SVM: Optimality of algorithms

We show here that the above SVM information complexity estimates are within a logarithmic term of being optimal. A simple heuristic argument would be as follows: for any nontrivial distribution $\rho \in F = \mathcal{P}$, standard random information results from Monte Carlo or IBC yield that for some non-constant function $g(x)$, with probability at least $1 - \delta$, the error between actual and empirical risk is

$$|R_\rho(g) - R_{\widehat{\rho}}(g)| = \Omega\big(1/\sqrt{n}\big). \qquad (4.14)$$

Indeed, even with a $g$ taking two values the above holds, and it easily follows that this holds for any $g$ which is essentially non-constant (i.e., is not equivalent to a constant function) on the support of $\rho$. An informal conclusion is that the above error of SVM in Theorem 4.1 is optimal within a log term.

A precise result along these lines is (Vapnik and Chernovenkis, 1974):

**Theorem 4.3 (Vapnik and Chernovenkis):** *If the function $L(g(\mathbf{x}), y)$ is essentially non-constant on the support of $\rho$, then for any $\delta > 0$, Theorem 4.1 fails to hold if the right hand side of (4.1.1) is replaced by any function $A(n) = o\left(\frac{1}{\sqrt{n}}\right)$.*

Thus the error in Theorem 4.1 is within a factor $\sqrt{\ln n}$ of being optimal, and it follows easily that the $\epsilon$-information compexity of SVM in Theorem 4.2 is within a logarithmic factor, $\ln\left(\frac{1}{\epsilon}\right)$, of being optimal.

# 5. SVM: Improvement of VC complexity bounds

We now show it is possible to improve the bounds in section 4 so as to eliminate the logarithmic term, if we restrict ourselves to a class of loss functions $L(g(\mathbf{x}), y)$ which are polynomial in the two arguments. Note this class is dense in the set of all continuous loss functions $L(g(\mathbf{x}), y)$ for compactly supported densities $\rho$.

## 5.1 Preliminaries

Recalling $G_1$ is the class of affine functions on $\mathbb{R}^d$, fix $M > 0$ and let $G_{1M}$ be the compact space of all affine $g(\mathbf{x}) = \mathbf{w} \cdot \mathbf{x} + b$ with $|\mathbf{w}|_1 \equiv \sum_i |w_i| < M$ and $b < M$. Letting $\rho \in F = \mathcal{P}$, with empirical information $N\rho = \{(\mathbf{x}_i, y_i)\}_{i=1}^d$, we first consider some probabilistic bounds. By Chebyshev's theorem

$$|R_{\widehat{\rho}}(g) - R(g)| = \left|\frac{1}{n}\sum_{i=1}^n L(g(\mathbf{x}_i), y_i) - E_\rho[L(g(\mathbf{x}), y)]\right| \leq \epsilon$$

with probability at least $1 - \frac{\mathcal{V}(L(g(\mathbf{x}),y))}{n\epsilon^2} \equiv 1 - \delta$ (where $\mathcal{V}$ is variance) or equivalently,

$$|R_{\widehat{\rho}}(g) - R(g)| \leq \sqrt{\frac{\mathcal{V}(L(g(\mathbf{x}), y))}{\delta n}} = \frac{\sigma(L(g, y))}{\sqrt{\delta n}} \tag{5.1}$$

with probability at least $1 - \delta$. This bound works for a single $g$, while our goal is to make this bound uniform over the class $g \in G_1$.

If $L(g(\mathbf{x}), y)$ is a polynomial in $g$ and $y$ (e.g., squared error loss $L(g(\mathbf{x}), y) = (g(\mathbf{x}) - y)^2$), we claim the corresponding "polynomial risk" SVM also carries the better bound (5.1) for the estimation of the risk by empirical risk uniformly over $g \in G_1$. Thus assume $L$

$$L(g(\mathbf{x}), y) = \sum_{0 \leq i+j \leq k} c_{ij} g^i y^j$$

is a polynomial of order $k$. Then the difference between risk and empirical risk (error) is

$$R_{\widehat{\rho}}(g) - R_\rho(g) = \sum_{0 \leq i+j \leq k} c_{ij}(E_{\widehat{\rho}} - E_\rho)(g^i y^j)$$

and since the difference is $O\left(\frac{1}{\sqrt{n}}\right)$ for each monomial in $\mathbf{x}$ above, we will show that this also holds uniformly in $g$ for the sum, initially by requiring that $g \in G_{1M}$. We first require a simple fact:

**Lemma 5.1:** *Given two functions $f$ and $g$ on a set $Z$ which take on their minima,*

$$\left| f\left( \arg\inf g \right) - \inf f \right| \leq 2 \sup_{z \in Z} |f(z) - g(z)|$$

*Proof:* Assume this does not hold. Letting $a = \sup_{z \in Z} |f(z) - g(z)|$, we would have

$$f(\arg\inf g) > \inf f + 2a.$$

This would imply

$$\inf g = g(\arg\inf g) \geq f(\arg\inf g) - a > \inf f + a.$$

We also have

$$\inf g \leq g(\arg\inf f) \leq \inf f + a,$$

which gives a contradiction. □

Note also that for any finite sum of functions $\sum_i f_i$, defining $\sigma(f) \equiv \sqrt{\mathcal{V}(f)}$, we have by the triangle inequality in $L^2(\rho)$,

$$\sigma\left(\sum_i c_i f_i\right) = \sigma\left(\sum_i c_i \overline{f}_i\right) = \left\|\sum_i c_i \overline{f}_i\right\|_2 \qquad (5.2)$$

$$\leq \sum_i \|c_i \overline{f}_i\|_2 = \sum_i |c_i| \sigma(f_i)$$

with $\overline{f} = f - E(f)$.

## 5.2 Polynomial risk functions

We now have:

**Lemma 5.2:** *For any $M > 0$ and any polynomial $L(g, y)$, we have with probability at least $1 - l\delta$, simultaneously for all $g \in G_{M1}$,*

$$|R_{\widehat{\rho}}(g) - R_\rho(g)| \leq \frac{\sum\limits_{0 \leq i+j \leq k} |c_{ij}| \sqrt{d_i} (2M)^i}{\sqrt{\delta n}}, \tag{5.3}$$

where $L(g(\mathbf{x}), y) = \sum\limits_{0 \leq i+j \leq k} c_{ij} g^i y^j$, $d_i \equiv E_\rho[(|\mathbf{x}|+1)^{2i}])$, and $l$ is the number of non-zero terms in $L(g, y)$ as a polynomial in $\mathbf{x} = (x_1, \ldots, x_d)$.

**Proof:** Writing $x_{d+1} = 1$, $w_{d+1} = b$ and defining $\mathbf{x}' = (x_1, \ldots, x_{d+1})$, $\mathbf{w}' = (w_1, \ldots, w_{d+1})$, we have for $L(g, y) = \sum\limits_{0 \leq i+j \leq k} c_{ij} g^i y^j$,

$$L(g, y) = L(\mathbf{w} \cdot \mathbf{x} + b, y) = L(\mathbf{w}' \cdot \mathbf{x}', y) = \sum_{0 \leq i+j \leq k} c_{ij}(\mathbf{w}' \cdot \mathbf{x}')^i y^j$$

$$= \sum_{0 \leq i+j \leq k} \pm c_{ij}(\mathbf{w}' \cdot \mathbf{x}')^i = \sum_{|\alpha| \leq k} \pm a_\alpha \mathbf{w}'^\alpha \mathbf{x}'^\alpha, \tag{5.4}$$

where the above sum is over all multiindices $\alpha = (\alpha_1, \ldots, \alpha_{d+1})$ with $\alpha_i$ non-negative integers and $|\alpha| = \sum\limits_i \alpha_i$. The total number of distinct powers $\mathbf{x}'^\alpha = x_1^{\alpha_1} \ldots x_d^{\alpha_d}$ in the last sum is $l$. Note that $l$ is the maximum number of distinct powers $\mathbf{x}^\alpha = x_1^{\alpha_1} \ldots x_d^{\alpha_d}$ which can appear in (5.4); in particular note the last term $x_{d+1}^{\alpha_{d+1}}$ in (5.4) is always 1.

Note

$$\sigma^2(f) = E(f^2) - E(f)^2 \leq E(f^2). \tag{5.5}$$

Let $\mathbf{w}_p$ be the vector defined by taking absolute values of components of $\mathbf{w}'$, i.e.,

$$\mathbf{w}_{pi} = |w'_i|$$

Then by (5.1) and (5.5), with probability at least $1 - l\delta$ (since we must use (5.1) $l$ times below)

$$|(R_{\widehat{\rho}} - R_\rho)(L(g, y))| = \left| \sum_{|\alpha| \leq k} \pm a_\alpha \mathbf{w}'^\alpha (R_{\widehat{\rho}} - R_\rho)(\mathbf{x}'^\alpha) \right|$$

$$\leq \frac{1}{\sqrt{\delta n}} \sum_{|\alpha| \leq k} |a_\alpha| |\mathbf{w}'^\alpha| \left( E(\mathbf{x}'^{2\alpha}) \right)^{1/2}$$

$$\leq \frac{1}{\sqrt{\delta n}} \sum_{i=0}^{k} \sum_{|\alpha|=i} |a_\alpha| |\mathbf{w}'^\alpha \mathbf{1}^\alpha| \left( E((1+|\mathbf{x}'|)^{2i}) \right)^{1/2} \qquad (5.6)$$

$$\leq \frac{1}{\sqrt{\delta n}} \sum_{0 \leq i+j \leq k} |c_{ij}| (\mathbf{w}_p \cdot \mathbf{1})^i \left( E((1+|\mathbf{x}'|)^{2i}) \right)^{1/2}$$

$$\leq \frac{1}{\sqrt{\delta n}} \sum_{0 \leq i+j \leq k} |c_{ij}| (2M)^i \sqrt{d_i}, \qquad (5.7)$$

with $\mathbf{1} = \begin{bmatrix} 1 \\ \vdots \\ 1 \end{bmatrix}$, since $\mathbf{w}_p \cdot \mathbf{1} = |\mathbf{w}'| = |\mathbf{w}| + |b| \leq 2M$. The next to last inequality above follows from the expansion (5.4). Note that if all $c_{ij} > 0$, this inequality is an equality for each fixed $i$ and thus also for the sum over $i$. The general argument is then not difficult, given that the components of $\mathbf{w}'$ which appear in both sums are in absolute value only.

We now need

**Lemma 5.3:** *For any probability distribution $\rho(\mathbf{x},y)$ on $\mathbb{R}^d \times \mathbb{R}$ and a polynomial loss $L(g,y)$ which is positive definite, the risk $R_\rho(g)$ attains its minimum over $g \in G_1$, with $G_1$ the class of affine functions. Further, there is no minimum at $\infty$, i.e., no minimizing sequence $g_i(\mathbf{x}) = \mathbf{w}_i \cdot \mathbf{x} + b_i$ with $\mathbf{w}_i$ or $b_i$ $\underset{i \to \infty}{\longrightarrow} \infty$, for which $R_\rho(g_i) \underset{i \to \infty}{\longrightarrow} \underset{g \in G_1}{\inf} R(g)$.*

**Proof:** We will assume there is no hyperplane (i.e., affine subspace) $H \subset \mathbb{R}^d$ on which $\rho(\mathbf{x},y)$ is supported in $\mathbf{x}$. For if such a hyperplane exists we can without loss restrict $\rho$ to it or a smaller hyperplane in which contains no proper sub-hyperplane on which $\rho$ is supported, which we assume has been done.

To prove the Lemma, let $g_i = \mathbf{w}_i \cdot \mathbf{x} + b_i$ be a minimizing sequence for $R_\rho$. We claim it suffices to show that $\mathbf{w}_i$ and $b_i$ must remain bounded. Indeed this would automatically prove the last statement of the Lemma. In addition, by taking subsequences, this would imply that $g_i$ converge pointwise to a fixed $g = \mathbf{w} \cdot \mathbf{x} + b$, and it is then easy to check that $g$ is a minimizer of $R_\rho$, showing that $R_\rho$ attains its minimum.

To show $\mathbf{w}_i$ and $b_i$ above remain bounded, assume this it is false for a contradiction. By the minimizing sequence assumption we have

$$R_\rho(g_i) \xrightarrow[i \to \infty]{} \inf_{g \in G_1} R_\rho(g).$$

On the other hand, since we assume $\mathbf{w}_i$ or $b_i$ are not bounded, either $\mathbf{w}_i$ or $b_i$ has a subsequence which converges to $\infty$. Assume first that a subsequence of $\mathbf{w}_i$ converges to $\infty$. By taking subsequences, assume $|\mathbf{w}_i| \xrightarrow[i \to \infty]{} \infty$. Then we claim $R_\rho(g_i) \xrightarrow[i \to \infty]{} \infty$ (independently of $b_i$). To show this, note that given $N > 0$ there is an $\alpha > 0$ such that $\{\mathbf{x} : |g(\mathbf{x})| = |\mathbf{w} \cdot \mathbf{x} + b| \leq N\}$ has measure smaller than $1 - \alpha$ (for all $b$) for $\mathbf{w}$ sufficienly large (since the width of this set, $\frac{N}{|\mathbf{w}|}$, becomes arbitrarily small, and $\rho$ is not supported on a proper hyperplane). Thus as $|\mathbf{w}_i| \to \infty$, $g(\mathbf{x})$ and hence $L(g, y)$ (which is positive definite) becomes arbitrarily large on a set of measure at least $\alpha$. Then we would have

$$R_\rho(g) = R_\rho(\mathbf{w}_i \cdot \mathbf{x} + b) \xrightarrow[i \to \infty]{} \infty,$$

which contradicts the assumption $R_\rho(g_i)$ converges.

Thus we must conclude $\{\mathbf{w}_i\}$ has no subsequence which converges to $\infty$, and $\{\mathbf{w}_i\}$ remains bounded. On the other hand, if a subsequence of $|b_i|$ converges to $\infty$, by taking a further subsequence we can have $\mathbf{w}_i \xrightarrow[i \to \infty]{} \mathbf{w}_0$ (since $\mathbf{w}_i$ are bounded) and $|b_i| \xrightarrow[i \to \infty]{} \infty$, which would imply $g_i(\mathbf{x}) = \mathbf{w}_i \cdot \mathbf{x} + b$ is not a minimizing sequence. Thus $\{b_i\}$ must also be bounded, completing the proof.

Now note by Theorem 4.1, the minimizer $\widehat{g} \in G_1$ of the empirical risk $R_{\widehat{\rho}}(g)$ is close to the minizer $g_1 \in G_1$ of the true risk $R_\rho(g)$, in that for any $\delta > 0$, with probability at least $1 - \delta$,

$$R(\widehat{g}) - \inf_{g \in G_1} R(g) = O\left(\sqrt{\frac{\ln n}{n}}\right). \tag{5.8}$$

By the last statement of Lemma 5.3, there are $M, \epsilon > 0$ such that $R(g) > \inf_{g_1 \in G_1}$

$R(g_1) + \epsilon$ if $g \notin G_{1M}$. Now choose an $M$ and $\epsilon$ as above. Now by (6.4), for $n$ sufficiently large $\widehat{g} \in G_{1M}$ (with probability at least $1 - \delta$), since eventually $R(\widehat{g}) - \inf_{g_1 \in G_1} R(g_1) < \epsilon$ with at least that probability. By Lemma 5.2, we have together with Lemma 5.1:

**Theorem 5.4:** *Given a probability measure $\rho(\mathbf{x}, y)$ on $\mathbb{R}^{d+1}$, a positive definite polynomial risk $L(g, y)$, a minimizer $\widehat{g}$ of the empirical risk is an $O(1/\sqrt{n})$ approximation to any true minimizer $g_1$ of the risk in $G_1$, in that for sufficiently large $n$, with probability at least $1 - l\delta$,*

$$R_\rho(\widehat{g}) - R_\rho(g_1) \leq 2 \frac{\sum_{0 \leq i+j \leq k} |c_{ij}| \sqrt{d_i}(2M)^i}{\sqrt{\delta n}},$$

*where $M$ is any constant (which always exists) such that all minimizers $g_1 \in G_1$ of $R_\rho(g)$ are in $G_{1M}$.*

Note the existence of a finite $M$ is guaranteed by the argument before the Theorem. We note that for a compactly supported $\rho$, we can approximate any continuous non-negative $L(g, y)$ uniformly by positive definite polynomials in $g$ and $y$ on the support of $\rho$, so we have

**Theorem 5.5:** *For a compactly supported $\rho$ and any continuous non-negative function $L(g, y) \geq 0$, there exists an $L^*$ which is arbitrarily close to $L$ (in sup-norm), such that asymptotic error (in the sense of risk) of an SVM using error criterion $L^*$ is of order $\frac{1}{\sqrt{n}}$, uniformly in $g \in G_1$.*

Note that the this result depends on $d_i$ and so is not uniform in $\rho$.

## 5.3 Uniform results in $\rho$

The above results arise from Lemma 5.2, which gives uniform bounds over $g \in G_1$. However, the bound in the Lemma is uniform in $\rho \in F = \mathcal{P}$ only for a class of $\rho$ for which $d_i = E_\rho[(|\mathbf{x}| + 1)^{2k}] < C$ for some fixed $C > 0$, where $k$ is the largest power of $g$ appearing in $L(g, y)$. This includes any class of $\rho$

supported in a fixed compact region in $\mathbb{R}^{d+1}$. In general, however, our bound is again local and not uniform in $\rho$, since polynomials are unbounded on $\mathbb{R}^d$.

In order to obtain uniform results in $\rho \in F$, we extend the above observations by noting that for the set of $L$ which are polynomial on a compact set $K$ in $(g, y) \in \mathbb{R}^2$ and constant outside $K$, the above results in fact become uniform in $\rho \in F$:

**Theorem 5.6 :** *If $L(g, y)$ is polynomial in $g$ and $y$ in any fixed compact set $K \subset \mathbb{R}^2$ of values $(g, y)$, and has constant value $C$ outside this set, then for all probability distributions $\rho$, with probability at least $1 - m\delta$,*

$$R_{\widehat{\rho}}(g) - R_{\rho}(g) \leq \frac{\sum_{1 \leq i+j \leq k} |c_{ij}| M_0^i + |c_{00} - C|/2}{\sqrt{\delta n}},$$

*where $M_0 = \sup_{(g,y) \in K} |g|$, and $m$ is the number of non-zero terms (in $g$ and $y$) in the polynomial $L(g, y)$.*

**Proof:** We have
$$L(g, y) = (L(g, y) - C) + C \equiv L_1(g, y) + C,$$

with

$$L_1 = L - C = \begin{cases} \sum_{0 \leq i+j \leq k} c_{ij} g^i y^j - C & \text{if } (g, y) \in K \\ 0 & \text{otherwise} \end{cases}.$$

Therefore with probability at least $1 - m\delta > 0$ (since there are at most $m$ distinct terms where $R_{\widehat{\rho}} - R_{\rho}$ is replaced by $\sigma$ below),

$$|R_{\widehat{\rho}}(g) - R_{\rho}(g)| \tag{5.9}$$

$$= \left|(R_{\widehat{\rho}} - R_{\rho})\left(\left[\sum_{0 \le i+j \le k} c_{ij} g^i y^j - C\right] \chi_K(g, y)\right)\right|$$

$$= \left|\sum_{0 \le i+j \le k} c_{ij}(R_{\widehat{\rho}} - R_{\rho})\left[g^i y^j \chi_K(g, y)\right] - C(R_{\widehat{\rho}} - R_{\rho})\chi_K(g, y)\right|$$

$$\le \frac{\sum_{1 \le i+j \le k} |c_{ij}|\sigma(g^i y^j \chi_K(g, y)) + |c_{00} - C|\sigma(\chi_K(g, y))}{\sqrt{\delta n}}.$$

Now we bound using (5.5):

$$\sigma^2(g^i y^j \chi_K) \le E(|g|^{2i}) \le M_0^{2i},$$

while

$$\sigma^2(\chi_K(g, y)) = E(\chi_K(g, y)) - E(\chi_K(g, y))^2 \le 1/4$$

since $E(\chi_K) \le 1$.

This gives

$$(R_{\widehat{\rho}}(g) - R_{\rho}(g)) \le \frac{\sum_{1 \le i+j \le k} |c_{ij}| M_0^i + |c_{00} - C|/2}{\sqrt{\delta n}},$$

a bound which is independent of $\rho \in F$. $\square$

Finally we have the uniform analog of Theorem 5.4:

**Theorem 5.7:** *Given a risk function L as in Theorem 5.4, a minimizer $\widehat{g}$ of empirical risk is an $O(1/\sqrt{n})$ approximation to any true minimizer $g_1$ of the risk $G_1$, uniformly in $\rho \in F = \mathcal{P}$. More precisely, for sufficiently large $n$, with probability at least $1 - m\delta$,*

$$R_{\rho}(\widehat{g}) - R_{\rho}(g) \le 2\frac{\sum_{0 \le i+j \le k} |c_{ij}| M_0^i + |c_{00} - C|/2}{\sqrt{\delta n}}.$$

This follows from Theorem 5.6 together with Lemma 5.1. Note since the bounds are uniform in $\rho \in F$, an argument such as in Lemma 5.3 is unnecessary here. □

## 6. Scaled families of algorithms

### 6.1 Uses of scaled families

Scaled families of algorithms can be useful because increased information typically can be used with increased algorithmic complexity. In some IBC applications increased algorithmic complexity is implcitly scaled with increased information complexity, as with spline algorithms, where more data points yield more spline knots. As a simple example of such scaling, note that given a large number (e.g., $10^6$) of data points, linear regression (with an approximation space $G$ consisting of affine functions) will generally under-utilize the data. One can enlarge the space of approximation algorithms to have a range made of approximating functions with more parameters, e.g., involving quadratics and cubics of the variables.

See Vapnik (1998, 2000) for an SLT analysis of such scalings of complexity, in which the VC dimension $h$ of the family of approximating functions (the range $G$ of allowable algorithms $\phi \in \Phi$) is scaled with cardinality of information. For information $N: F \to \mathbb{R}^n$ of cardinality $n$, we can formalize such a scaling by choosing an algorithm $\phi_k: \mathbb{R}^n \to G_k$ whose range $G_k$ has VC dimension $h(k)$, with the scaling $h = h(k)$ chosen so that the error of approximation is minimized.

Defining $g_0$ to be the true regression function, i.e., a minimizer in the full space $G \supset \bigcup_n G_n$ of the risk $R(g)$, again define (assuming the arg inf exist)

$$g_k = \arg\inf_{g \in G_k} |R(g) - R(g_0)| = \text{closest element in } G_k \text{ to } g_0$$

$$\widehat{g}_k = \arg\inf_{g \in G_k} (R_{\widehat{\rho}}(g)) = \text{minimizer in } G_k \text{ of empirical risk with } n \text{ data points.}$$

Recalling the definitions in Section 2.2 we note informally that $e_{\inf}(n, k)$ decreases as $n \to \infty$, and $e_{\text{alg}}(k)$, which can be bounded in terms of VC dimension $h(k)$ of $G_k$, goes to 0 as $h \to \infty$. If $h$ is too large relative to $n$, we have overfitting: estimation (information) error becomes large, as we are in the wrong space (with too many functions or parameters). In this case the goal is to decrease $h = h(k)$ in order to lower $e_{\text{inf}}$. In general we scale the VC dimension $h = h(k)$ of $\Phi_k$ (all algorithms with range $G_k$) with the information complexity $n$. We want the number of free parameters, measured by $h$ (related to the algorithmic complexity), scaled to data cardinality $n$ (information complexity). This approach is taken in Kon and Plaskota (2000), where algorithms are a scaled family of neural networks (with algorithmic complexity defined as the number of neurons).

## 6.2 Scaling of algorithms: nonlinear SVM generalizations

**Scaled families of algorithms:** By Theorem 4.1 the information error of an SVM is asymptotically bounded (with probability at least $1 - \delta$) as

$$e_{\text{inf}}(n, k) \equiv \epsilon \leq K\sqrt{\mathcal{E}} = K\sqrt{4\frac{h(\ln\frac{2n}{h} + 1) - \ln(\frac{\delta}{8})}{n}}$$

with $h = h(k) = \text{VC}(G_k)$, the VC dimension of the class $G_k$ of SVM decision functions $g(\mathbf{x})$, and $K > 0$. Making the above discussion more precise, we see that to guarantee $e_{\text{inf}}(n, k)$ vanishes as $n \to \infty$, we need a scaling of $n$ with $k$ so that $h(k)/n$ is decreasing (Vapnik 1998, 2000). On the other hand, $h(k)$ must increase with $n$ so that algorithmic error $e_{\text{alg}}(k) \xrightarrow[k\to\infty]{} 0$. Thus $\phi_k$ is defined by

$$\widehat{g}_n = \phi_k(N_n(\rho)) \in G_k = \text{Ran}(\phi_k)$$

and $k$ is chosen so $h(k) = \text{VC}(G_k)$, the dimension of the range of $\phi_k$, scales with $n$.

**An SVM example:** To give an example of this, let $P_k$ be the polynomials of degree $k$ on $\mathbb{R}^d$. The standard SVM algorithm (minimizing the loss function $L(g(\mathbf{x}), y) = (1 - g(\mathbf{x})y)_+$) will be denoted as

$$\phi_1 : \mathbf{Z} \equiv \{\mathbf{z}_i\}_{i=1}^n \to G_1 = P_1.$$

The space of loss functions $H = \{L(p(\mathbf{x}), y)\}_{p \in P_1}$ (as functions of $\mathbf{x}$ and $y$) has VC dimension $\text{VC}(H) \leq d + 2$, as shown in section 4.3.

We now extend the standard space $\Phi_1$ of SVM algorithms with range $P_1$ to a scaled set $\{\Phi_k\}_{k=1}^{\infty}$, with $\Phi_k$ the set of algorithms with range in $P_k$. Define the range $G_k$ of such algorithms for $k > 1$ to be *nonlinear SVM*. Practically such algorithms can be implemented by extending the data vector $N\rho = \mathbf{x} = (x_1, \ldots, x_n) = \{x_i\}_{i=1}^n$ to

$$\tilde{\mathbf{x}} = \{\mathbf{x}^\alpha\}_{|\alpha| \leq k}, \tag{6.1}$$

consisting of all monomials

$$\mathbf{x}^\alpha = x_1^{\alpha_1} \ldots x_n^{\alpha_n}$$

of degree $k$ or less in components $x_i$, and then using a standard SVM algorithm (with range in the affine polynomials $P_1$ in $\tilde{\mathbf{x}}$). Above $\alpha = (\alpha_1, \ldots, \alpha_n)$, with $\alpha_i$ non-negative integers and $|\alpha| = \sum_{i=1}^n \alpha_i \leq k$.

The dimension of $P_k$ is the number of monomials in $d$ variables of order less than $k$, i.e., the cardinality of $\{\alpha : |\alpha| \leq k\}$. This is the number of non-negative lattice points $\alpha = (\alpha_1, \ldots, \alpha_n)$ in $d$ dimensions satisfying $\sum_i \alpha_i \leq k$, which is of order of the volume of the region $\left\{\sum_{i=1}^d x_i \leq k\right\}$ in the positive octant of $\mathbb{R}^d$. Thus

$$D(k) = \Theta(k^d) \qquad (k \to \infty)$$

grows polynomially in $k$, where $c_1 k^d \leq \Theta(k^d) \leq c_2 k^d$.

To scale $k$ with $n$, note that the VC dimension of $P_k$ is $h(k) \leq D(k)$, since the VC dimension of a space of functions is bounded by the dimension of its non-constant part plus 1 (see Korian and Sontag, 1997).

Thus by Theorem 4.1, recalling (2.5), we have

**Theorem 6.1:** *For the above scaled family of nonlinear (polynomial) SVM algorithms, the δ-probabilistic error (error with probability at least $1 - \delta$) satisfies*

$$e(n, k) \leq \frac{J_k \tau_k a(p) \sqrt{\mathcal{E}}}{\left(1 - \tau_k a(p) \sqrt{\mathcal{E}}\right)_+} + O\left(\frac{1}{n}\right) + e_{\text{alg}}(k), \tag{6.2}$$

where $\mathcal{E} = 4\frac{h(\ln\frac{2n}{h}+1) - \ln(\frac{\delta}{8})}{n}$, $J_k = \inf_{g \in P_k} R(g)$, $h = h(k) \leq D(k)$ is the VC dimension of the polynomial space $P_k$, and $\tau_k$ is as in (4.1), with $G_1$ replaced by $G_k$.

The algorithmic (approximation) error

$$e_{\text{alg}}(k) = \inf_{g \in P_k} |R(g) - R(g_0)|,$$

$$= \inf_{g \in P_k} \left| \int_{\mathbb{R}^{n+1}} L(g(\mathbf{x}), y) - L(g_0(\mathbf{x}), y) d\rho(\mathbf{x}, y) \right|, \tag{6.3}$$

is an approximation theory measurement of the distance between the optimal $g_0 \in G \supset \bigcup_k G_k$ (the full function space) and its approximation by polynomials $g_k \in P_k$, using the error (6.3). This can be bounded analytically (see the next section) or through simulations.

## 6.3 Bounding the algorithmic error

We wish to scale $k$ (which determines algorithmic error and hence algorithmic complexity) with information complexity $n$ by letting the range of our algorithms vary through a the scale $G_k \subset G$, with $k = 1$ for standard SVM. To have an error bound which decreases, we can choose $G_k$ so as to scale $h = h(k) \leq D(k) = \Theta(k^d)$ to grow more slowly than $n$, e.g., $k = o(n^{1/d})$.

If we wish to scale $k$ with $n$ to minimize the right side of (6.2), there is a scaling prescription based on bounds on the full error $e(n, k)$. As an example of this, if the a priori distribution $\rho \in F$ is assumed supported in $\mathbf{x}$ on the ball $B_r(0) \subset \mathbb{R}^d$ and is assumed to admit a risk-minimizing function $g_0 = \arg\inf_{g' \in G} R(g)$ all of whose (multiple) directional derivatives $(\widehat{\mathbf{v}} \cdot D)^k \equiv D_{\widehat{\mathbf{v}}}^k$ (in all unit directions $\widehat{\mathbf{v}}$ in $\mathbf{x}$) are bounded by a constant $A$ (for $|\mathbf{z}| \leq r$), then Taylor's theorem with remainder in the direction from $\mathbf{z} = 0$ to $\mathbf{z} = \mathbf{z}_1$ gives for $g = g_0$:

$$g(\mathbf{z}_1) = \sum_{k=0}^{K} \frac{(\mathbf{z}_1 \cdot D)^k g(0)}{k!} + \frac{(\mathbf{z}_1 \cdot D)^{K+1} g(\mathbf{z}_2)}{(K+1)!} \tag{6.4}$$

where $\mathbf{z}_2$ is in the line between 0 and $\mathbf{z}_1$, and $D$ acts only on the $\mathbf{z}$ variable which is the argument of $g$, after which it is evaluated at 0 in the first sum.

We can bound the error in (6.4) as

$$\left| \frac{(\mathbf{z}_1 \cdot D)^{K+1} g(\mathbf{z}_2)}{(K+1)!} \right| = \frac{1}{(K+1)!} |\mathbf{z}_1|^{K+1} |(D_{\widehat{\mathbf{z}_1}} g)^{K+1}(\mathbf{z}_2)|$$

$$\leq \frac{A}{(K+1)!} |r|^{K+1}.$$

Therefore (letting $g_k = \arg\inf_{g \in G_k} R(g)$):

$$e_{\text{alg}}(k) = R(g_k) - R(g_0) = \int_{\mathbb{R}^{d+1}} [(1 - g_k y)_+ - (1 - g_0 y)_+] \, d\rho(\mathbf{x}, y),$$

$$\leq \|g_k y - g_0 y\|_\infty = \|g_k - g_0\|_\infty$$

$$\leq \frac{A}{(K+1)!} r^{K+1}$$

since $g_k \in G_k$ can be chosen as the $k^{th}$ order Taylor polynomial approximation to $g_0$ in $P_k$ and the minimum risk can be bounded by the risk of this $g_k$.

In this example a choice $k = k(n)$ can be made which minimizes the right side of (6.2) above. Note that such an optimization (in this case on the sum of our upper bounds) occurs when the terms $e_{\text{alg}}(k)$ and

$$e_{\inf}(n, k) \leq 2 J_k \tau_k a(p) \sqrt{4 \frac{h(\ln \frac{2n}{h} + 1) - \ln(\frac{\delta}{8})}{n}}$$

(which holds for $n$ sufficiently large) have rates of change with respect to $k$ which are of the same order, which gives a minimum in $k$.

As mentioned above, dimension reduction (e.g., a projection $P(\mathbf{x})$ of $\mathbf{x}$) is useful for pruning the set of possible $g(\mathbf{x})$ as approximations to the unknown $\rho \in F = \mathcal{P}$. This is important when the dimension of range spaces $G_k$ of algorithms $\phi_k \in \Phi_k$ grows rapidly, e.g., where $G_k = P_k$ is the set of multivariate polynomials above. In the latter case the feature (information) vector

$$N(\rho) = \{\tilde{\mathbf{x}}_i\}_{1 \leq i \leq n} = \{\mathbf{x}_i^\alpha\}_{1 \leq i \leq n;\, |\alpha| \leq k}, \qquad \text{where} \qquad \mathbf{x}_i^\alpha \equiv \prod_{j=1}^d x_{ij}^{\alpha_j}, \qquad \text{with}$$

$\mathbf{x}_i = (x_{i1}, \ldots, x_{id})$. Dimension reduction can be done either through elimination of less relevant variables $\mathbf{x}^\alpha$ or by pruning coordinates from the basic vector $\mathbf{x}$.

## 6.4 An example of scaling of algorithmic and information complexity

**The Gaussian case:** For SVM, recall the use of scaled families of nonlinearized feature vectors $N(\rho)$ and corresponding approximation algortithms $\phi_k$ is a *nonlinear* SVM (NLSVM). Whether there is advantage to using NLSVM reduces in a Gaussian situation to the question: given two multivariate Gaussian distributions

$$\rho_1(\mathbf{x}) \equiv \frac{1}{(2\pi)^{d/2} \det \Sigma_1} e^{-\frac{1}{2}(\mathbf{x}-\mu_1)^T \Sigma_1^{-1}(\mathbf{x}-\mu_1)} = N(\mu_1, \Sigma_1), \tag{6.5}$$

$$\rho_2(\mathbf{x}) \equiv N(\mu_2, \Sigma_2) \tag{6.6}$$

(the conditional distributions $\rho_1 = \rho(\mathbf{x}|y=1)$ and $\rho_2 = \rho(\mathbf{x}|y=-1)$ of the feature vectors conditioned on $y = 1$ and $y = -1$, respectively), what is the shape of an optimal separator between the two? With this assumption that $\rho_i$ are Gaussian, our goal here is to form an SVM separator $g(\mathbf{x})$ for which the risk function

$$R(g) = P_1(g(\mathbf{x}) < 0) + P_2(g(\mathbf{x}) > 0)$$

(proportional the expected number of errors) is minimized. In addition, weighting of false positives versus false negatives may also be useful, in which case the new risk function is

$$R_1(g) = \beta_1 P_1(g < 0) + \beta_2 P_2(g > 0), \tag{6.7}$$

with $\beta_1 + \beta_2 = 1$. We may be less concerned with false positives than false negatives, so $\beta_1 \gg \beta_2$ is a good choice, if the overall numbers of positive examples are larger than numbers of negative examples.

We consider the complexity of the SVM with risk function (6.7), focusing on the algorithmic (approximation) error. If our algorithm $\phi$ based on data $N\rho = \mathbf{Z} = \{(\mathbf{x}_i, y_i)\}_{i=1}$ restricts $g$ to the class $G_1$ of affine functions and if algorithmic error is large, a decision to extend $G_1$ to a nonlinear SVM instead makes sense. It is then of interest to find the approximation error for hypothesis spaces $G_k$ which include polynomials of order $k$, with $k$ increasing.

**Finding the optimal solution:** For the risk function (6.7) we can in fact identify the optimal choice of $g$ among *all* functions (not just affine ones), if we first show the separation surface

$$g(\mathbf{x}) = \beta_1 \rho_1(\mathbf{x}) - \beta_2 \rho_2(\mathbf{x}) = 0, \tag{6.8}$$

(for the above choices (6.5, 6.6) of $\rho_i$) is optimal - this can be done using calculus of variations. Indeed, if there is an infinitesimal variation in the optimal $g$ resulting in a change $dV$ of the volume separated by $g(\mathbf{x}) = 0$, in the direction $g(\mathbf{x}) > 0$ at location $\mathbf{x}$, then the first order increment in $R_1(g)$ above is

$$dR_1(g) = \beta_1 \rho_1(\mathbf{x}) dV - \beta_2 \rho_2(\mathbf{x}) dV = 0,$$

since we are increasing the volume in which $g < 0$ and decreasing that in which $g > 0$; we have $dR_1 = 0$ since $R_1(g)$ is stationary.

To higher than first order it follows easily that since $\beta_1 \rho_1(\mathbf{x})$ is on the average larger than $\beta_2 \rho_2(\mathbf{x})$ in the volume $dV$ (since $\rho_1(\mathbf{x})$ increases in the direction $g(\mathbf{x}) > 0$ from $g(\mathbf{x}) = 0$), the risk (1) has increased. Symmetry shows that when $dV$ is in the opposite direction, $R_1$ increases as well, proving the optimizing surface is (6.8). The above balance between $\beta_i \rho_i(\mathbf{x})$ in the direction $g(\mathbf{x}) > 0$ follows more directly by noting that from (6.5)

$$\ln \beta_i \rho_i = -\frac{1}{2}(\mathbf{x} - \boldsymbol{\mu}_i)^T \Sigma_i^{-1} (\mathbf{x} - \boldsymbol{\mu}_i) + C_i. \tag{6.9}$$

where $C_i$ is a constant.

Thus the surface $g(\mathbf{x}) = 0$ is given by equality of two quadratic polynomials of the form (6.9), namely by

$$-\ln \beta_1 + \frac{1}{2}\ln \det \Sigma_1 + (\mathbf{x} - \mu_1)^T \Sigma_1^{-1}(\mathbf{x} - \mu_1),$$

$$= -\ln \beta_2 + \frac{1}{2}\ln \det \Sigma_2 + (\mathbf{x} - \mu_2)^T \Sigma_2^{-1}(\mathbf{x} - \mu_2). \tag{6.10}$$

In general (if $\Sigma_1 \neq \Sigma_2$) this surface is quadratic, so that use of a quadratic SVM (in which $G = G_2 = P_2$) is appropriate. From this we expect that in general cases where the distributions of positive and negative classes have different covariances $\Sigma_1 \neq \Sigma_2$, so that the quadratic terms in (6.10) do not cancel, there may be significant improvement using a quadratic SVM (giving a quadratic separation surface) over a linear one. This is illustrated in the example of the Wisconsin cancer study below.

However, inequality of covariances $\Sigma_i$ for positive and negative examples is not always the case. An example involves data in computational biology for which the improvement from linear to quadratic SVM is marginal (Cvetkovski, et al.), implying that in such cases positives and negatives have distributions (if approximately Gaussian) with about the same covariances.

In case of the risk function (6.7), the surface $g = 0$ is (from (6.10))

$$\mathbf{x}^T\left(\Sigma_1^{-1} - \Sigma_2^{-1}\right)\mathbf{x} - 2\left(\Sigma_1^{-1}\mu_1 - \Sigma_2^{-1}\mu_2\right)^T\mathbf{x} + \mu_1^T\Sigma_1^{-1}\mu_1$$

$$- \mu_2^T\Sigma_2^{-1}\mu_2 + \frac{1}{2}\ln\left(\frac{\beta_2^2\det\Sigma_1}{\beta_1^2\det\Sigma_2}\right) = 0.$$

The effect of the weighting coefficients $\beta_i$ is to shift the surface without changing its shape. The size of $\Sigma_1^{-1} - \Sigma_2^{-1}$ determines whether the quadratic SVM will improve risk significantly over the linear one. This suggests a general criterion for appropriateness of a quadratic SVM - given empirical covariances $\widehat{\Sigma}_1$ and $\widehat{\Sigma}_2$ of the two data sets (assumed sufficiently large), the norm $\|\Sigma_1^{-1} - \Sigma_2^{-1}\|$ should be small.

## 6.5 Example: Application to biomedical informatic data

We apply here the above example of a scaled algorithm family to some data in biomedical informatics, the Wisconsin cancer database (Radwin, 1992). We begin with a standard SVM applied to 9 input variables (measured physical characteristics of a tumor), which predict the output variable $y$, which is cancer malignancy (+1) or non-malignancy (−1). The data, summarized in the table below, are taken from a random selection of 349 training examples and 349 test examples out of 699 total data. The first test via SVM (with data involving all 9 input variables) has an error rate of 13.75% on the test set. When a dimensional reduction is done and the three most useful variables are extracted, there is an SVM error rate of 32.39%. When the nonlinear SVM of degree 2 is applied to these input data, the total error rate goes down to 8.60%.

| Machine \ Error rate | FP | FN | TP | TN | ERR | %ERR |
|---|---|---|---|---|---|---|
| 9-variable SVM | 37 | 11 | 107 | 194 | 48 | .1375 |
| 3-variable SVM | 41 | 72 | 44 | 192 | 113 | .3239 |
| 3-variable nonlinear SVM | 29 | 1 | 117 | 202 | 30 | .0860 |

Table: F/TP = false/true positive; F/TN = false/true negative; ERR-total errors

This significant improvement of the quadratic over the linear SVM implies that the $3 \times 3$ covariance matrices $\Sigma_1$ and $\Sigma_2$ for the 3-variable data are significantly different between positive and negative examples. There are some current analogous methods (Holloway, et al., 2006 for the linear case) for identifying transcription initiation sites in the genome from examples, using several SVM feature spaces.

## References

Braverman, M. and S. Cook (2006). Computing over the reals: Foundations for scientific computing. *Notices AMS* **53**, 318-329.

Blum, L, F. Cucker, M. Shub, and S. Smale (1998). *Complexity and Real Computation.* Springer, New York.

Cvetkovski, A., C. DeLisi, D. Holloway, M. Kon, and P. Seal (2006). Dimensional reduction and optimization in TF-gene binding inferences. Technical report, Boston University.

Holloway, D., M. Kon and C. DeLisi (2006). Machine learning for regulatory analysis and transcription factor target prediction in yeast. Preprint, to appear, *Systems and Synthetic Biology.*

Radwin, M. (1992). Wisconsin breast cancer database,

http://www.radwin.org/michael/projects/learning/about-breast-cancer-wisconsin.html

Kon, M. and L. Plaskota (2000), Information complexity of neural networks. *Neural Networks* **13**, 365-376.

Kon, M. and L. Raphael (2006), Statistical learning theory and uniform approximation bounds in wavelet spaces, preprint.

Koiran, P and E. Sontag (1997). Vapnik-Chervonenkis dimension of recurrent neural networks. *Proceedings of Third European Conference on Computational Learning Theory*, Jerusalem.

Traub, J., G. Wasilkowski, and H. Woźniakowski (1988). *Information-Based Complexity.* Academic Press, Boston.

Traub, J. and A. Werschulz (1998). *Complexity and Information.* Cambridge University Press, Cambridge.

Vapnik, V. (1998). *Statistical Learning Theory.* Wiley, New York 1998.

Vapnik, V. (2000). *The Nature of Statistical Learning Theory.* Springer, New York, 2000.

Vapnik, V. and A. Chernovenkis (1974). *Theory of Pattern Recognition.* Nauka, Moscow.

Zhou, D. and S. Smale (2003). Estimating the approximation error in learning theory. *Analysis and Applications* **1**, 1-25.